\documentclass[twoside,11pt]{article}

\usepackage{automl2020}

\usepackage{euler}
\usepackage{amsfonts}       
\usepackage{nicefrac}       
\usepackage{microtype}      

\usepackage{color}
\usepackage{amsopn}
\usepackage{amsmath}
\usepackage{amssymb}
\usepackage{empheq}
\usepackage{hyperref}

\usepackage{adjustbox}
\usepackage{algorithm,algorithmicx}
\usepackage{algpseudocode}
\usepackage[small]{caption}

\usepackage{subcaption}
\usepackage{todonotes}
\usepackage[normalem]{ulem}

\makeatletter
\newcommand{\leqnomode}{\tagsleft@true\let\veqno\@@leqno}
\newcommand{\reqnomode}{\tagsleft@false\let\veqno\@@eqno}
\makeatother

\newcommand{\argmin}{\operatornamewithlimits{arg\,min}}

\DeclareMathOperator*{\st}{\text{subject to}}
\DeclareMathAlphabet\mathbfcal{OMS}{cmsy}{b}{n}
\newcommand{\Def}[0]{\mathrel{\mathop:}=}
\newcommand{\bz}{\mathbf z}
\newcommand{\bu}{\mathbf u}
\newcommand{\btheta}{\boldsymbol{\theta}}
\newcommand{\tbtheta}{\widetilde{\boldsymbol{\theta}}}
\newcommand{\bdelta}{\boldsymbol{\delta}}
\newcommand{\blambda}{\boldsymbol{\lambda}}
\newcommand{\iter}[2]{#1{}^{(#2)}}

\jmlrheading{%
P. Ram, S. Liu, D. Vijaykeerthi, D. Wang, D. Bouneffouf, G. Bramble,
H. Samulowitz and A. G. Gray%
}

\ShortHeadings{Constrained CASH with ADMM}{%
Ram, Liu, Vijaykeerthi, Wang, Bouneffouf, Bramble, Samulowitz, and Gray%
}
\firstpageno{1}

\begin{document}
\title{Solving Constrained CASH Problems with ADMM}
\author{%
  \name P. Ram \email p.ram@gatech.edu \\
  \name S. Liu \email sijia.liu@ibm.com \\
  \name D. Vijaykeerthi \email deepakvij@in.ibm.com \\
  \name D. Wang \email dakuo.wang@ibm.com \\
  \name D. Bouneffouf \email djallel.bouneffouf1@ibm.com \\
  \name G. Bramble \email gbramble@us.ibm.com \\
  \name H. Samulowitz \email samulowitz@us.ibm.com \\
  \name A. G. Gray \email alexander.gray@ibm.com \\
  \addr IBM Research
}
\maketitle
\begin{abstract}
The CASH problem has been widely studied in the context of automated
configurations of machine learning (ML) pipelines and various solvers and
toolkits are available. However, CASH solvers do not directly handle black-box
constraints such as fairness, robustness or other domain-specific custom
constraints. We present our recent approach~\citep{liu2020admm} that leverages
the ADMM optimization framework to decompose CASH into multiple small problems
and demonstrate how ADMM facilitates incorporation of black-box constraints.
\end{abstract}
\section{Automated ML Pipeline Configuration} \label{sec:intro}
Hyper-parameter optimization (HPO) for a {\em single} machine learning (ML)
algorithm is widely studied in AutoML~\citep{snoek2012practical,
  shahriari2016taking}. HPO was generalized to the {\bf C}ombined {\bf
  A}lgorithm {\bf S}election and {\bf H}PO (CASH) problem to configure multiple
stages of a ML pipeline (transformers, feature selectors, predictive models)
automatically~\citep{autoweka1,feurer2015efficient}. Since the CASH formulation,
various innovative solvers have been
proposed~\citep{smac,bergstra2011algorithms,mohr2018ml,rakotoarison2019automated}.
CASH has two main challenges: (i) the tight coupling between the algorithm
selection \& HPO; and (ii) the black-box nature of optimization objective
lacking any explicit gradients -- feedback is only available in the form of
(often expensive) function evaluations. Furthermore, the CASH formulation does
not explicitly handle constraints on the individual ML pipelines
such as fairness or domain-specific constraints on individual ML pipelines.
%

We view CASH as a mixed integer black-box nonlinear program and we recently
proposed a novel solution framework leveraging the {\em alternating direction
  method of multipliers} (ADMM) framework~\citep{liu2020admm}. ADMM offers a
{\em two-block} alternating optimization procedure that splits an involved
problem (with multiple variables \& constraints) into simpler (often
unconstrained) sub-problems \citep{boyd2011distributed,parikh2014proximal}.

\noindent
\textbf{Contributions.}
We utilize ADMM to decompose CASH into 3 problems: (i) a black-box optimization
with a {\em small} set of {\em only continuous} variables, (ii) a closed-form
Euclidean projection onto an integer set, and (iii) a black-box integer
program. Moreover, the ADMM framework handles {\em any black-box constraints
  alongside the black-box objective} -- such constraints are seamlessly
incorporated while retaining very similar sub-problems.
\subsection{Related work} \label{sec:litreview}
Beyond grid-search and random search \citep{bergstra2012random} for HPO,
sequential model-based optimization (SMBO) is a common technique with different
`surrogate models' such as Gaussian processes~\citep{snoek2012practical}, random
forests \citep{smac} and tree-parzen estimators~\citep{bergstra2011algorithms}.
Successive halving \citep{jamieson2016non,sabharwal2016selecting} and HyperBand
\citep{li2018hyperband} utilize computationally cheap {\em multi-fidelity}
approximations of the objective based on some budget (training samples/epochs)
with bandit learning to eliminate unpromising candidates early. These schemes
essentially perform an efficient random search over discrete or
discretized continuous spaces. BOHB~\citep{bohb} combines SMBO (with TPE) and
HyperBand. Meta-learning~\citep{vanschoren2018meta, fusi2018probabilistic,
  drori2018alphad3m} leverages past experiences with search space refinements
and promising starting points.
%
SMBO with a large number of variables along with conditional dependencies has
been used to solve CASH in the widely-used Auto-WEKA~\citep{autoweka1,autoweka2}
and Auto-sklearn~\citep{feurer2015efficient} toolkits. Both apply the general
purpose SMAC framework~\citep{smac} to find optimal ML pipelines of fixed
shape.
Hyperopt-sklearn~\citep{komer2014hyperopt} utilizes TPE as the SMBO.

SMBO-based CASH has been improved by partially splitting CASH and taking
advantage of the structure in the algorithm selection
problem. ML-Plan~\citep{mohr2018ml} uses hierarchical task networks (HTN)
planning for algorithm selection and randomized search for HPO, while
MOSAIC~\citep{rakotoarison2019automated} utilizes Monte-Carlo Tree Search (MCTS)
and Bayesian Optimization (BO) respectively. The two sub-problems are coupled
with a shared surrogate model. Even though algorithm selection and HPO are
solved in independent steps, MOSAIC still requires computations (such as the
acquisition function estimation/optimization and the surrogate-model training)
over the high-dimensional joint search-space of all algorithms and
hyper-parameters (HPs). Our proposed ADMM-based solution performs an explicit
primal-dual decomposition, significantly reducing the dimensionality of each of
the aforementioned sub-problems. Moreover, our solution presents a general
framework that can incorporate HTN-Planning and MCTS based algorithm selection.

%
ADMMBO~\citep{ariafar2019admmbo} for general BO with black-box constraints
maintains a separate surrogate model for the objective and each of the black-box
constraints (a total of $M+1$ surrogate models for $M$ constraints) and utilizes
the ADMM framework to split the constrained optimization into a sequence of
unconstrained problems to minimize the objective and constraint violations. The
sub-problem dimensionality remains the same as the original problem. Our
proposed ADMM based scheme for CASH with black-box constraints takes advantage
of the problem structure to handle $M$ black-box constraints by adding $M$
variables to one of the three sub-problems without needing any additional
surrogate models.
\section{CASH as a Mixed Integer Nonlinear Program}\label{sec:prob-setting}
We focus on CASH for a {\em fixed pipeline shape} -- for $N$ \textit{functional
  modules} with $K_i$ choices in each module, let $\bz_i \in \{0, 1\}^{K_i}$
denote the algorithm choice in module $i$, with the constraint $\mathbf{1}^\top
\bz_i = 1$ ensuring a single choice per module.  Let $\bz = \left\{ \bz_1,
\ldots, \bz_N \right\}$. Assuming that categorical HPs can be
encoded as integers, let $\btheta_{ij}$ be the HPs of
algorithm $j$ in module $i$ ($\btheta_{ij}^c \in \mathcal{C}_{ij} \subset
\mathbb{R}^{m_{ij}^c}$ \textit{continuous} and $\btheta_{ij}^d \in
\mathcal{D}_{ij} \subset \mathbb{Z}^{m_{ij}^d} $ \textit{integer}).  With
$\btheta^c = \left\{ \btheta_{ij}^c, \forall i \in [N], j \in [K_i] \right\}$,
and $\btheta^d$ defined analogously, let $f(\bz, \{\btheta^c, \btheta^d\})$
represent the loss of a ML pipeline configured with algorithm choices $\bz$ and
the HPs $\{\btheta^c, \btheta^d\}$. CASH can be stated
as\footnotemark:
{\small
\begin{equation}\label{eq:prob0}\tag{CASH}
  \displaystyle \min_{\bz, \btheta^c, \btheta^d} f(\bz, \{\btheta^c,
  \btheta^d\}) \st \left\{
  \begin{array}{l}
    \bz_i \in \{0,1 \}^{K_i}, \mathbf 1^{\top} \bz_i = 1, \forall i \in [N],\\
    \btheta_{ij}^c \in \mathcal{C}_{ij}, \btheta_{ij}^d \in \mathcal{D}_{ij},
    \forall i \in [ N ], j \in [ K_i ].
  \end{array}
  \right.
\end{equation}
}
\footnotetext{ The above formulation allows for easily extending the search to
  more flexible pipelines (Appendix~\ref{asec:gen-pipelines}). To extensively
  evaluate our proposed CASH solver with multiple repetitions and restarts, we
  also propose a novel cheap-to-evaluate black-box objective
  (Appendix~\ref{asec:art-bbopt}) that possesses the structure of
  \eqref{eq:prob0}.}
%
We may need the ML pipelines to also {\em explicitly satisfy} application
specific {\em constraints} corresponding to $M$ general {\em black-box
  functions} with no analytic form in $(\bz, \{\btheta^c, \btheta^d\})$:
{\small
\begin{equation} \label{eq:prob0-csts}
  g_m(\bz, \{\btheta^c, \btheta^d\}) \leq \epsilon_m, m \in [M].
\end{equation}
}
For example, deployment constraints may require prediction latency below a
threshold. Business constraints may require highly accurate pipelines with explicitly bounded false positive rate -- false positives may
deny loans to eligible applicants, which is lost business and could violate
anti-discriminatory requirements. Pursuing {\em fair} AI, regulators may require
that any deployed ML pipeline explicitly satisfies bias constraints
\citep{friedler2019comparative}.
\section{Decomposing CASH with ADMM} \label{sec:ADMM}
Introducing a surrogate objective $\widetilde{f}$ over the continuous domain
$(\mathcal{C}_{ij} \times \widetilde{\mathcal{D}}_{ij}), i \in [N], j \in [K_i]$
that matches the objective $f$ over $(\mathcal{C}_{ij}\times \mathcal{D}_{ij})$
with $\widetilde{\mathcal{D}}_{ij}$ as the continuous relaxation of the integer
space $\mathcal{D}_{ij}$, we follow ADMM, detailed in
Appendix~\ref{asec:admm-decomp}, to decompose \eqref{eq:prob0} into the
following 3 sub-problems to be solved iteratively, with $(t)$ representing the
ADMM iteration index, $\rho$ as the penalty for the augmented Lagrangian term
and the Lagrangian multipliers $\blambda$ updated as $\iter{\blambda}{t+1} =
\iter{\blambda}{t} + \rho (\iter{\tbtheta^d}{t+1} - \iter{\bdelta}{t+1} )$:
{\small
\begin{align}
  \label{eq:theta-min} \tag{$\btheta$-min} \left\{
    \iter{\btheta^c}{t+1}, \iter{\tbtheta^d}{t+1}
  \right\} & = \argmin_{
    \btheta^c_{ij}, 
    \tbtheta^d_{ij} 
  } \widetilde{f} \left(
    \iter{\bz}{t},  \left\{\btheta^c, \tbtheta^d \right\}
  \right)
  + \frac{\rho}{2} \left\| \tbtheta^d - \mathbf{b} \right\|_2^2,
    \mathbf{b} \Def
    \iter{\bdelta}{t} - \frac{1}{\rho} \iter{\blambda}{t},\\
  \label{eq:delta-min} \tag{$\bdelta$-min} \iter{\bdelta}{t+1}
    & = \argmin_{\bdelta_{ij} \in \mathcal{D}_{ij}}
      \left\| \mathbf{a} - \bdelta \right\|_2^2, \quad
    \mathbf{a} \Def \iter{\tbtheta^d}{t+1} + (1 / \rho) \iter{\blambda}{t},\\
  \label{eq:z-step} \iter{\bz}{t+1} & =
    \argmin_{\bz_i \in \{0,1 \}^{K_i}, \mathbf1^{\top} \bz_i = 1}
    \widetilde{f} \left(
    \bz, \left\{\iter{\btheta^c}{t+1}, \iter{\tbtheta^d}{t+1} \right\}
    \right).
    \tag{$\bz$-min}
\end{align}
}

\noindent
\textbf{Solving~\eqref{eq:theta-min}.}
This is a continuous black-box optimization problem with
variables $\btheta^c$ and $\tbtheta^d$. Since the algorithms $\iter{\bz}{t}$ are
fixed in \eqref{eq:theta-min}, $\widetilde{f}$ only depends on the
HPs of the chosen algorithms -- the {\em active variable set} $S = \{(\btheta_{ij}^c, \tbtheta_{ij}^d): \iter{z_{ij}}{t} = 1
\}$. This splits \eqref{eq:theta-min} even further into (i) $
\min_{\tbtheta_{ij}^d \in \widetilde{\mathcal{D}}_{ij}} \| \tbtheta_{ij}^d -
\mathbf{b}_{ij} \|_2^2$ with $z_{ij} = 0$ (the {\em inactive} set) requiring
a Euclidean projection of $\mathbf{b}_{ij}$ onto $\widetilde{\mathcal{D}}_{ij}$,
and (ii) a black-box optimization with a {\em small} active continuous variable
set\footnotemark $S$.
%
\footnotetext{For the CASH problems we consider in our empirical evalutations, $
  | \btheta | = | \btheta_{ij}^c | + | \tbtheta_{ij}^d | \approx 100 $ while the
  largest possible active set $S$ is less than $15$ and typically less than
  $10$.}
Solvers such as BO \citep{shahriari2016taking}, direct
search \citep{larson2019derivative}, or trust-region based derivative-free
optimization \citep{conn2009introduction} work fairly well.

\noindent
\textbf{Solving~\eqref{eq:delta-min}.}
This requires an elementwise minimization $\min_{\bdelta_{ij}} (\bdelta_{ij} -
\mathbf{a}_{ij})^2$ and solved in closed form by projecting $\mathbf{a}_{ij}$
onto $\widetilde{\mathcal{D}}_{ij}$ and then rounding to the nearest integer in
$\mathcal{D}_{ij}$.

\noindent
\textbf{Solving~\ref{eq:z-step}.}
The following black-box discrete optimizers can be utilized:
(i)~{\it MCTS} can be used as in \citet{rakotoarison2019automated}, (ii)~{\em
  HTN-Planning} as in \citet{mohr2018ml} (iii)~\textit{Multi-fidelity
  evaluations} can be used with successive halving
\citep{jamieson2016non,li2018hyperband} or incremental data allocation
\citep{sabharwal2016selecting}, (iv)~Interpreting \eqref{eq:z-step} as a
\textit{combinatorial multi-armed bandits} problem and utilizing Thompson
sampling \citep{durand2014thompson} as we did in \citet[Appendix
  4]{liu2020admm}.
\subsection{Empirical Advantages of ADMM}
ADMM has been used for both convex~\citep{boyd2011distributed} and non-convex
optimization~\citep{xu2016empirical} and various ADMM enhancements have been
proposed. We demonstrate how these improve ADMM for \eqref{eq:prob0}. For
initial evaluations, we use the artificial objective
(Appendix~\ref{asec:art-bbopt}) for cheap evaluations, which allow us to
efficiently generate statistically significant results over multiple trials. A
maximum of 100 ADMM iterations is executed with $\rho=1$ (ADMM for CASH appears
to be stable with respect to $\rho$; see Appendix~\ref{asec:ADMM-rho-sense}) for
a maximum runtime of $2^{10}$ seconds. We consider a search space of 4 modules
with 8 scalers, 11 transformers, 7 feature selectors and 11 estimators (total
6776 algorithm combinations with almost $100$ HPs). See \citet[Appendix
  7]{liu2020admm} for complete details.
\begin{figure}[tb]
  \centering
  \begin{subfigure}{0.233\textwidth}
    \includegraphics[width=\textwidth]{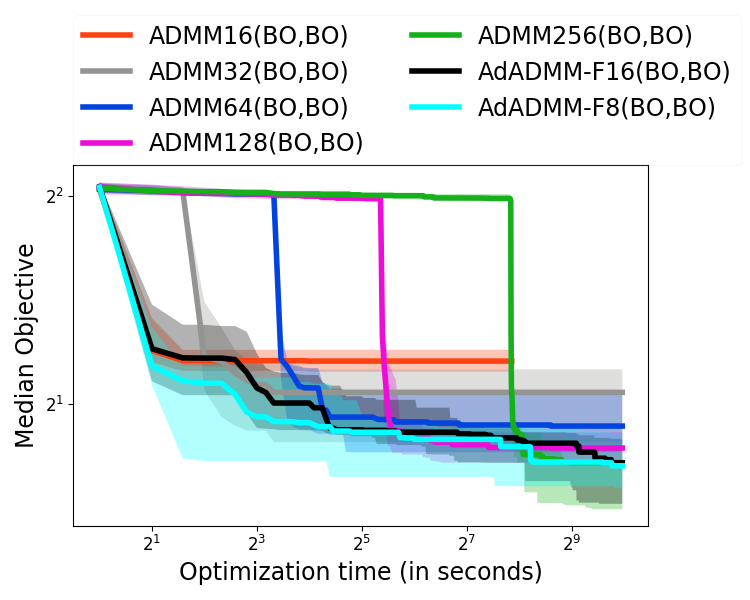}
    \caption{Fixed vs. Adaptive}
    \label{fig:admm-mods:fixed-v-adaptive}
  \end{subfigure}
  ~
  \begin{subfigure}{0.233\textwidth}
    \includegraphics[width=\textwidth]{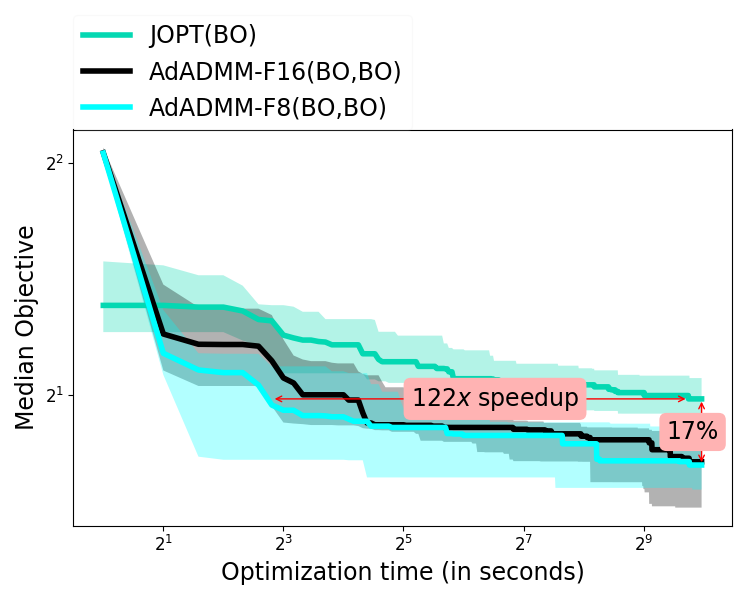}
    \caption{Split vs. Joint}
    \label{fig:admm-mods:split-v-joint}
  \end{subfigure}
  ~
  \begin{subfigure}{0.233\textwidth}
    \includegraphics[width=\textwidth]{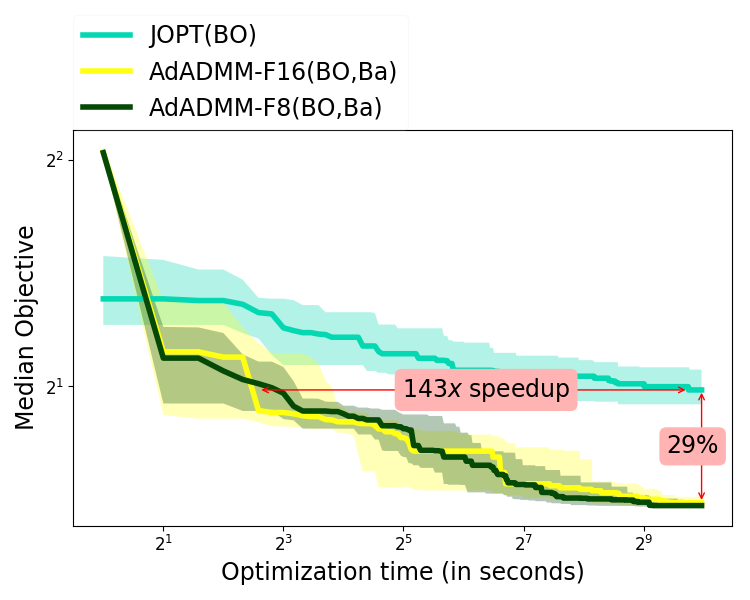}
    \caption{Custom solvers}
    \label{fig:admm-mods:solvers}
  \end{subfigure}
  ~
  \begin{subfigure}{0.233\textwidth}
    \includegraphics[width=\textwidth]{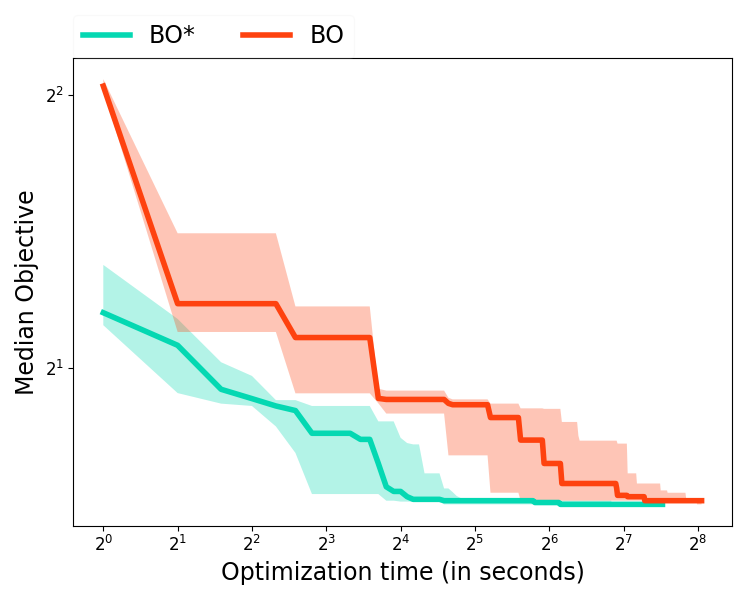}
    \caption{Cold vs. Warm}
    \label{fig:admm-mods:warm}
  \end{subfigure}
  \caption{Performance of ADMM and enhancements on the artificial black-box
    objective (Appendix~\ref{asec:art-bbopt}) with the incumbent objective
    (median and inter-quartile range over 30 trials) on the vertical axis and
    optimization time on the horizontal axis. Note the logscale on both axes.}
  \label{fig:admm-mods}
\vskip -0.15in
\end{figure}

\noindent
\textbf{Adaptive precision.}
ADMM progresses by iteratively solving \eqref{eq:theta-min} and
\eqref{eq:z-step} (\eqref{eq:delta-min} is solved in closed form). In practice,
the sub-problems are often solved to lower precision in the initial ADMM
iterations; the precision is progressively increased through the ADMM iterations
({\em adaptive precision}) instead of maintaining the same precision for all
ADMM iterations ({\em fixed precision}). In
Figure~\ref{fig:admm-mods:fixed-v-adaptive}, we use Bayesian Optimization (BO)
to solve both \eqref{eq:theta-min} \& \eqref{eq:z-step}. The
precision of the sub-problems are controlled by modifying the number of BO
iterations. For fixed precision ADMM, we fix the BO iterations for the
sub-problems to $I = \{16, 32, 64, 128, 256 \}$ denoted as ADMM$16$(BO,BO) and
so on. For adaptive precision ADMM, we initially solve each sub-problem with
$16$ BO iterations and progressively increase the number of BO iterations to
$256$ with an additive factor of $F = \{8, 16\}$ in each ADMM iteration denoted
by AdADMM-$F8$(BO,BO) \& AdADMM-$F16$(BO,BO) respectively. We see the expected
behavior -- fixed precision ADMM with small $I$ dominate for small time scales
but saturate soon; large $I$ require significant start-up time but dominate for
larger time scales. Adaptive precision ADMM provides best anytime performance.
%

\noindent
\textbf{Solving smaller sub-problems.}  In
Figure~\ref{fig:admm-mods:split-v-joint}, we demonstrate the advantage of
solving multiple smaller sub-problems over solving a single large joint
problem. We use BO to solve both the joint problem (JOPT(BO)) and the
sub-problems in the adaptive ADMM (AdADMM-$F8$(BO,BO) \&
AdADMM-$F16$(BO,BO)). ADMM demonstrates $120\times$ speedup to reach the best
objective obtained by JOPT(BO) within the time budget, and also provides an
additional $17\%$ improvement over that objective at the end of the time budget.

\noindent
\textbf{Custom sub-problem solvers.}  BO is designed for problems with (a small
number of) continuous variables and hence is well-suited for
\eqref{eq:theta-min}; it is not designed for \eqref{eq:z-step}. We should
instead use schemes customized for \eqref{eq:z-step} such as
MCTS~\citep{rakotoarison2019automated} or Thompson sampling for combinatorial
multi-armed bandits~\citep[Appendix 4]{liu2020admm}. In
Figure~\ref{fig:admm-mods:solvers}, we consider BO for \eqref{eq:theta-min} and
bandits for \eqref{eq:z-step} -- AdADMM-$F8$(BO,Ba) \& AdADMM-$F16$(BO,Ba) --
which further improves over adaptive ADMM, demonstrating $140\times$
speedup over JOPT(BO) with an additional $29\%$ improvement in the objective
value.

\noindent
\textbf{Warm Start ADMM.}
It is common in ADMM to {\em warm-start} the sub-problem minimizations in any
ADMM iteration with the solution of the same sub-problem from previous ADMM
iterations (if available) to improve empirical convergence. In adaptive
precision ADMM, we employ warm-starts for BO in \eqref{eq:theta-min} to get
BO$^*$ and compare it to BO with cold-starts. Figure~\ref{fig:admm-mods:warm}
indicates that while BO and BO$^*$ converge to the same objective at the end of
the time budget, BO$^*$ has significantly better anytime performance.

\noindent
\textbf{Evaluation on OpenML data sets.}
We evaluate the performance of ADMM against JOPT(BO) for \eqref{eq:prob0} on 8
OpenML data sets in Appendix~\ref{asec:eval-admm-jopt} and see over $10\times$
speedup in most cases and over $10\%$ improvement in the final objective in many
cases. We refer readers to \citet{liu2020admm} for the detailed comparison of
ADMM to Auto-sklearn~\citep{feurer2015efficient} and TPOT~\citep{olson2016tpot}
across 30 classification data sets.
\section{CASH with Black-box Constraints} \label{sec:ADMM:csts}
We consider \eqref{eq:prob0} in the presence of black-box constraints
\eqref{eq:prob0-csts}.  Without loss of generality, let $\epsilon_m \geq 0$ for
$m \in [M]$.  With scalars $u_m \in [0, \epsilon_m]$, we reformulate the
inequality constraint \eqref{eq:prob0-csts} as an equality constraint 
$g_m (\bz, \{ \btheta^c, \btheta^d \}) = \epsilon_m - u_m$ and a box
constraint $u_m \in [0, \epsilon_m]$.

Introducing (i) surrogate $\widetilde{g}_m$ for each black-box function $g_m, m
\in [M]$ over the continuous domain in a manner similar to $\widetilde{f}$ in
Section~\ref{sec:ADMM}, (ii) new Lagrangian multipliers $\mu_m, m \in [M]$ for
each of the $M$ black-box constraints, and following the ADMM mechanics
(detailed in Appendix~\ref{asec:ADMM:csts}), we decompose constrained
\eqref{eq:prob0} into the following unconstrained problems:
{\small
\begin{equation} \label{eq:cbbc-s-0} 
  \min_{\btheta^c_{ij}, \tbtheta^d_{ij}, u_m}
      \widetilde{f} ( \iter{\bz}{t},  \{\btheta^c, \tbtheta^d 
      \} ) 
      + \frac{\rho}{2} [
      \| \tbtheta^d - \mathbf{b} \|_2^2 + \sum_{m=1}^{M} [
        \widetilde{g}_m ( \iter{\bz}{t},  \{\btheta^c, \tbtheta^d
         \} ) - \epsilon_m + u_m  +
        \frac{1}{\rho}\iter{\mu_m}{t} ]^2 ],
\end{equation}
\begin{equation} \label{eq:ibbc-sz-0}
    \min_{\bz}
    \widetilde{f} (\bz,  \{\iter{\btheta^c}{t+1}, 
      \iter{\tbtheta^d}{t+1} \} )
    + \frac{\rho}{2} \sum_{i=1}^{M} [
      \widetilde{g}_m ( \bz, \{\iter{\btheta^c}{t+1},
      \iter{\tbtheta^d}{t+1} \}
      ) - \epsilon_m + \iter{u_m}{t+1}  + \frac{1}{\rho}
      \iter{\mu_m}{t}  ]^2,
\end{equation}
}
with $\mathbf{b}$ defined in \eqref{eq:theta-min}, $\bdelta$ updated as per
\eqref{eq:delta-min}, and $\mu_m, m \in [M]$ updated as $\iter{\mu_m}{t+1} =
\iter{\mu_m}{t} + \rho ( \widetilde{g}_m ( \iter{\bz}{t+1},
\{\iter{\btheta^c}{t+1}, \iter{\tbtheta^d}{t+1} \} ) - \epsilon_m +
\iter{u_m}{t+1} )$.

Problem~\eqref{eq:cbbc-s-0} is black-box optimization with continuous variables
similar to \eqref{eq:theta-min} (further split into active and inactive set of
variables as per $\iter{\bz}{t}$). The main difference from \eqref{eq:theta-min}
are the $M$ new optimization variables $u_m, m \in [M]$ active in every ADMM
iteration. This active set optimization can be solved in the same way as in
\eqref{eq:theta-min} (namely with BO).
Problem~\eqref{eq:ibbc-sz-0} remains a black-box integer program in $\bz$ with
an updated objective incorporating constraint violations and can be solved with
methods discussed for \eqref{eq:z-step}.
\subsection{Empirical evaluation}
\begin{figure}[tb]
  \centering
  \begin{subfigure}{0.235\textwidth}
    \includegraphics[width=\textwidth]{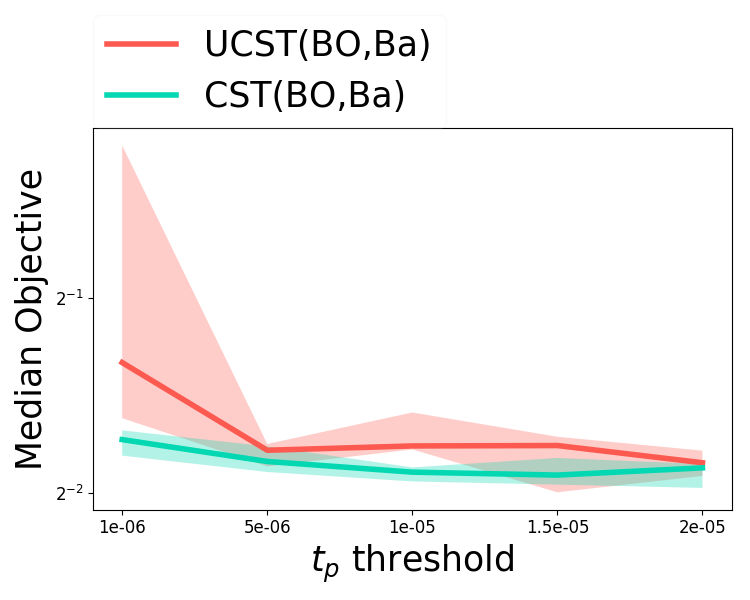}
    \caption{Objective vs. $t_p$}
    \label{fig:obj-t_p}
  \end{subfigure}
  ~
  \begin{subfigure}{0.235\textwidth}
    \includegraphics[width=\textwidth]{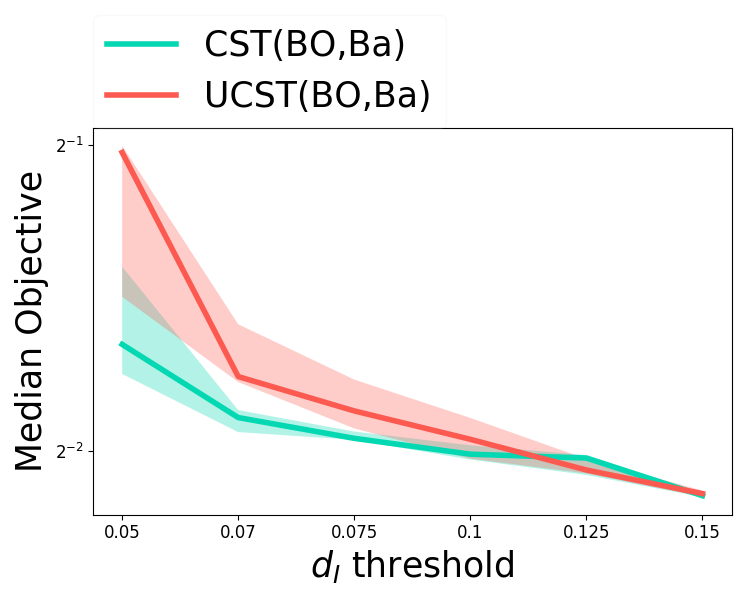}
    \caption{Objective vs. $d_I$}
    \label{fig:obj-d_i}
  \end{subfigure}
  \begin{subfigure}{0.235\textwidth}
    \includegraphics[width=\textwidth]{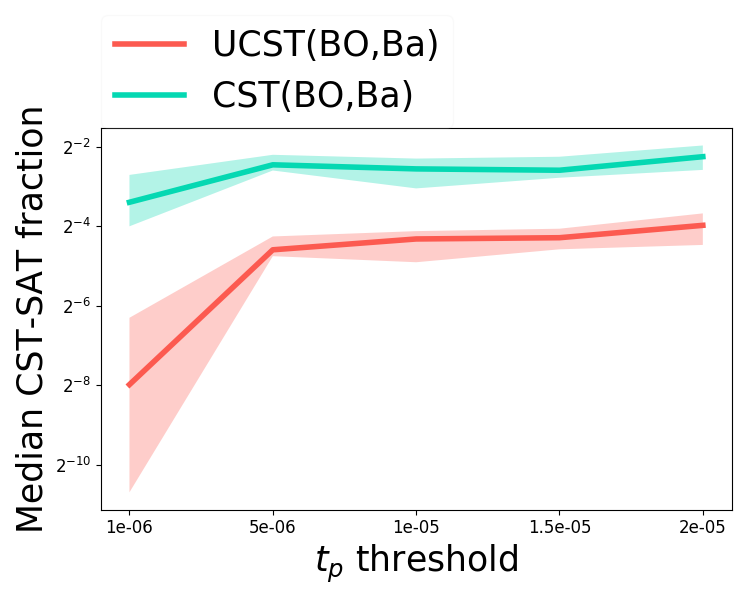}
    \caption{CST-SAT vs. $t_p$}
    \label{fig:csat-t_p}
  \end{subfigure}
  ~
  \begin{subfigure}{0.235\textwidth}
    \includegraphics[width=\textwidth]{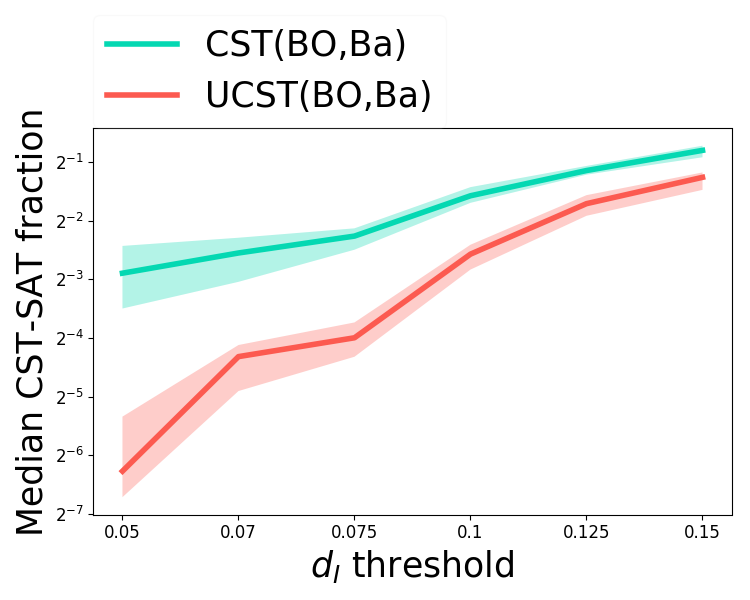}
    \caption{CST-SAT vs. $d_I$}
    \label{fig:csat-d_i}
  \end{subfigure}
  \caption{Performance of unconstrained and constrained ADMM (aggregated over 10
    trials) executed for 1 hour with varying thresholds for the 2
    constraints. Note the log-scale on the vertical axis.}
  \label{fig:bbo-bbc}
\vskip -0.15in
\end{figure}
We consider data from the Home Credit Default Risk Kaggle challenge with the
objective of $(1 - \mbox{AUROC})$, and 2 black-box constraints: (i) ({\bf
  deployment}) Prediction latency $t_p$ enforcing real-time predictions, (ii)
({\bf fairness}) Maximum pairwise disparate impact $d_I$
\citep{calders2010three} across all loan applicant age groups enforcing fairness
across groups.
We perform the following experiments: (i) fixing $d_I = 0.7$, we vary $t_p \in
[1, 20]$ (in $\mu s$), and (ii) fixing $t_p = 10\mu s$ and we vary
$d_I \in [0.05, 0.15]$. Note that the constraints get less
restrictive as the thresholds increase. We apply ADMM to the unconstrained
problem (UCST) and post-hoc filter constraint satisfying pipelines to
demonstrate that these constraints are not trivially satisfied.  Then we execute
ADMM with these constraints (CST).  Using BO for \eqref{eq:theta-min} (and
\eqref{eq:cbbc-s-0}) \& bandits for \eqref{eq:z-step} (and
\eqref{eq:ibbc-sz-0}), we get UCST(BO,Ba) \& CST(BO,Ba).

Figures~\ref{fig:obj-t_p} \& \ref{fig:obj-d_i} present the best objective
achieved when limited only to constraint satisfying pipelines
as the constraint on $t_p$ and $d_I$ are respectively relaxed. As expected, the
objective improves as the constraints relax. In both cases, CST outperforms
UCST, with UCST approaching CST as the constraints relax. Figures
\ref{fig:csat-t_p} \& \ref{fig:csat-d_i} (for varying $t_p$ \& $d_I$
respectively) present the constraint satisfying capability of the optimizer by
considering the fraction of constraint-satisfying pipelines found. CST again
significantly outperforms UCST, indicating that the constraints are non-trivial
to satisfy, and that ADMM is able to effectively incorporate the constraints for improved performance.
\section{Conclusion} \label{sec:conc}
In this paper, we summarized our recent ADMM based CASH solver and demonstrated
the utility of ADMM and its multiple enhancements. We also considered
CASH with black-box constraints and demonstrated how ADMM seamlessly handles them. While we skip the
multi-fidelity aspect of CASH here, ADMM can
use multi-fidelity solvers such as
Hyperband and BOHB for the
\eqref{eq:theta-min} and \eqref{eq:z-step} sub-problems. Moreover, we can combine ADMM with a context-free grammar for ML pipelines and AI planning
to jointly solve CASH and the pipeline shape search problem with black-box
constraints~\citep{katz2020exploring}.
\vskip 0.2in
{\small
\bibliography{ref}
}
\newpage
\appendix
\section{Generalization of the CASH Problem \eqref{eq:prob0}} \label{asec:gen-pipelines}

\paragraph{Generalization for more flexible pipelines.}
We can extend the problem formulation \eqref{eq:prob0} to enable optimization
over the ordering of the functional modules. For example, we can choose between
`preprocessor $\to$ transformer $\to$ feature selector' OR `feature selector
$\to$ preprocessor $\to$ transformer'.  The ordering of $T \leq N$ modules can
be optimized by introducing $T^2$ Boolean variables $\mathbf{o} = \{ o_{ik}
\colon i,k \in [T] \}$, where $o_{ik} = 1$ indicates that module $i$ is placed
at position $k$.  The following constraints are then needed:

\begin{enumerate}
\item[(i)] $\sum_{k \in [T]} o_{ik} = 1, \forall i \in [T]$, indicating that
  module $i$ is placed at a single position, and
\item[(ii)] $\sum_{i \in [T]} o_{ik} = 1 \forall k \in [T]$ enforcing that only
  one module is placed at position $k$.
\end{enumerate}

These variables and constraints can be added to $\bz$ in problem
\eqref{eq:prob0} ($\bz = \{ \bz_1, \ldots, \bz_N, \mathbf{o} \}$). The resulting
formulation still obeys the generic form of \eqref{eq:prob0}, which as we will
demonstrate in the following section, can be efficiently solved by an operator
splitting framework like ADMM.  We can also extend the above formulation to
allow the choice of multiple algorithms from the same module. Note that we
increase the combinatorial complexity of the problem as we extend the
formulation.

\section{Artificial black-box objective} \label{asec:art-bbopt}


We want to devise an artificial black-box objective to study the behaviour of
the proposed schemes and baselines that matches the properties of the AutoML
problem \eqref{eq:prob0} where

\begin{enumerate}
    \item The same pipeline (the same algorithm choices $\bz$ and the same
      hyperparameters $\btheta$) always gets the same value.
    \item The objective is not convex and possibly non-continuous.
    \item The objective captures the conditional dependence between $\bz_i$ and
      $\btheta_{ij}$ -- the objective is only dependent on the hyper-parameters
      $\btheta_{ij}$ if the corresponding $z_{ij} = 1$.
    \item Minor changes in the hyper-parameters $\btheta_{ij}$ can cause only
      small changes in the objective.
    \item The output of module $i$ is dependent on its input from module $i-1$.
\end{enumerate}

\paragraph{Novel artificial black-box objective.}
To this end, we propose the following novel black-box objective that emulates
the structure of \eqref{eq:prob0}:
\begin{itemize}
    \item For each $(i,j), i \in [N], j \in [K_i]$, we fix a weight vector
      $\mathbf{w}_{ij}$ (each entry is a sample from $\mathcal{N}(0,1)$) and a
      seed $s_{ij}$.
    \item We set $f_0 = 0$.
    \begin{itemize}
        \item For each module $i$, we generate a value
        $$v_i = \sum_{j} z_{ij} \left| \frac{\mathbf{w}_{ij}^\top
          \btheta_{ij}}{\mathbf{1}^T \btheta_{ij}} \right|$$ which only depends
          on the $\btheta_{ij}$ corresponding to the $z_{ij} = 1$, and the
          denominator ensures that the number (or range) of the hyper-parameters
          does not bias the objective towards (or away from) any particular
          algorithm.
        \item We generate $n$ samples $\{ f_{i,1}, \ldots, f_{i,n} \} \sim
          \mathcal{N}(f_{i-1}, v_i)$ with the fixed seed $s_{ij}$, ensuring that
          the same value will be produced for the same pipeline.
        \item $f_i = \max\limits_{m = 1, \ldots, n} | f_{i,m} |$.
    \end{itemize}
    \item Output $f_N$
\end{itemize}
The basic idea behind this objective is that, for each operator, we create a
random (but fixed) weight vector $\mathbf{w}_{ij}$ and take a weighted
normalized sum of the hyper-parameters $\btheta_{ij}$ and use this sum as the
scale to sample from a normal distribution (with a fixed seed $s_{ij}$) and pick
the maximum absolute of $n$ (say 10) samples. For the first module in the
pipeline, the mean of the distribution is $f_0 = 0.0$. For the subsequent
modules $i$ in the pipeline, the mean $f_{i-1}$ is the output of the previous
module $i-1$. This function possesses all the aforementioned properties of the
AutoML problem \eqref{eq:prob0}.

In black-box optimization with this objective, the black-box evaluations are
very cheap in contrast to the actual AutoML problem where the black-box
evaluation requires a significant computational effort (and hence
time). However, we utilize this artificial objective to evaluate the black box
optimization schemes when the computational costs are dominated by the actual
derivative-free optimization and not by the black-box evaluations.

\section{ADMM Decomposition for \eqref{eq:prob0}} \label{asec:admm-decomp} 

\paragraph{Introduction of continuous surrogate loss.}
We begin by proposing a surrogate loss of problem \eqref{eq:prob0}, which can
be defined over the continuous domain.  With $\widetilde{\mathcal{D}}_{ij}$ as
the continuous relaxation of the integer space $\mathcal{D}_{ij}$ (if
$\mathcal{D}_{ij}$ includes integers ranging from $\{l, \ldots, u \}\subset
\mathbb{Z}$, then $\widetilde{\mathcal{D}}_{ij} = [ l, u ] \subset \mathbb{R}$),
and $\tbtheta^d$ as the continuous surrogates for $\btheta^d$ with
$\tbtheta_{ij} \in \widetilde{\mathcal{D}}_{ij}$ (corresponding to $\btheta_{ij}
\in \mathcal{D}_{ij}$), we utilize a surrogate loss function $\widetilde{f}$ for
problem \eqref{eq:prob0} defined solely over the continuous domain with
respect to $\btheta$:

{\small
\begin{equation} \label{aeq:bbloss-continuous-surrogate}
\widetilde{f} \left(
  \bz,  \left\{\btheta^c, \tbtheta^d \right\}
\right) \Def f\left(
  \bz,  \left\{\btheta^c, \mathcal{P}_{\mathcal{D}} \left( \tbtheta^d \right)
\right\} \right),
\end{equation}
}

where
$\mathcal{P}_{\mathcal{D}} ( \tbtheta^d ) = \{ \mathcal{P}_{\mathcal{D}_{ij}} (
   \tbtheta_{ij}^d ), \forall i \in [N], j \in [K_i] \}$
is the projection of the continuous surrogates onto the integer set.  This
projection is {\bf necessary} since the black-box function is defined (hence
{\em can only be evaluated}) on the integer sets $\mathcal{D}_{ij}$s, not the
relaxed continuous set $\widetilde{\mathcal{D}}_{ij}$s. Note that this
projection and has an efficient closed form.  Given the above definitions, we
have the following \textit{equivalent} form of the problem \eqref{eq:prob0}:

{\small
\begin{equation}\label{aeq:prob0-rr}
  \displaystyle \min_{\bz, \btheta^c, \tbtheta^d, \bdelta}
  \widetilde{f} \left(
    \bz,  \left\{\btheta^c, \tbtheta^d \right\}
  \right) \st
  \left\{
    \begin{array}{l}
      \bz_i \in \{0,1 \}^{K_i}, \mathbf 1^{\top} \bz_i = 1, \forall i \in [N] \\
      \btheta_{ij}^c \in \mathcal{C}_{ij},
        \tbtheta_{ij}^d \in \widetilde{\mathcal{D}}_{ij},
        \forall i \in [ N ], j \in [ K_i ] \\
      \bdelta_{ij} \in \mathcal{D}_{ij}, \forall i \in [ N ], j \in [ K_i ] \\
      \tbtheta_{ij}^d  = \bdelta_{ij}, \forall i \in [ N ], j \in [ K_i ],
    \end{array}
  \right.
\end{equation}
}

where the equivalence between problems \eqref{eq:prob0} \&
\eqref{aeq:prob0-rr} is established by the equality constraint
$\tbtheta_{ij}^d = \bdelta_{ij} \in \mathcal{D}_{ij}$, implying
$\mathcal{P}_{\mathcal{D}_{ij}} ( \tbtheta_{ij}^d )
  = \tbtheta_{ij}^d \in \mathcal{D}_{ij} $
and $\widetilde{f} (\bz, \{\btheta^c, \tbtheta^d \} ) =
  f(\bz, \{\btheta^c, \tbtheta^d \} )$,
thereby making the objective functions in problems \eqref{eq:prob0} and
\eqref{aeq:prob0-rr} equal.  We highlight that the introduction of the continuous
surrogate loss \eqref{aeq:bbloss-continuous-surrogate} is the key to efficiently
handling AutoML problems \eqref{eq:prob0} by allowing us to perform
theoretically grounded operator splitting methods, e.g., ADMM, over mixed
continuous/integer hyper-parameters and integer model selection variables.

\paragraph{Operator splitting from ADMM.}
Using the notation that $I_\mathcal{X}(\mathbf{x}) = 0$ if
$\mathbf{x} \in \mathcal{X}$ else  $+\infty$, and defining the sets
{\small
\begin{align}
  \label{aeq:zcsets} \mathcal{Z} & = \left\{
    \bz \colon \bz = \{ \bz_i \colon \bz_i \in \{0,1 \}^{K_i},
    \mathbf 1^{\top} \bz_i = 1, \forall i \in [N] \}
  \right\}, \\
  \label{aeq:ccsets} \mathcal{C} & = \left\{
    \btheta^c \colon \btheta^c = \{ \btheta_{ij}^c \in \mathcal{C}_{ij}
    \forall i \in [N], j \in [K_i] \}
  \right\}, \\
  \label{aeq:dcsets} \mathcal{D} & = \left\{
    \bdelta \colon \bdelta = \{ \bdelta \in \mathcal{D}_{ij}
    \forall i \in [N], j \in [K_i] \}
  \right\}, \\
  \label{aeq:tdcsets} \widetilde{\mathcal{D}} & = \left\{
    \tbtheta^d \colon \tbtheta^d = \{
    \tbtheta_{ij}^d \in \widetilde{\mathcal{D}}_{ij}
    \forall i \in [N], j \in [K_i] \}
  \right\},
\end{align}
}
we can re-write problem \eqref{aeq:prob0-rr} as
{\small
\begin{align} \label{aeq:prob1}
  \displaystyle   \min_{\bz, \btheta^c, \tbtheta^d, \bdelta} \widetilde{f}
  \left(\bz,  \left\{\btheta^c, \tbtheta^d \right\} \right)
  + I_\mathcal{Z}(\bz) + I_\mathcal{C}(\btheta^c)
  + I_{\widetilde{\mathcal{D}}}(\tbtheta^d) + I_\mathcal{D}(\bdelta)
  \st\ \  \tbtheta^d = \bdelta,
\end{align}
}
with the corresponding augmented Lagrangian function
{\small
\begin{align}
  \label{aeq:prob1-al}
  \mathcal{L} & ( \bz, \btheta^c, \tbtheta^d, \bdelta, \blambda ) \Def \\
  \nonumber
  & \widetilde{f} \left(
    \bz,  \left\{\btheta^c, \tbtheta^d \right\}
  \right) + I_\mathcal{Z}(\bz) + I_\mathcal{C}(\btheta^c)
  + I_{\widetilde{\mathcal{D}}}(\tbtheta^d) + I_\mathcal{D}(\bdelta)
  + \blambda^\top \left( \tbtheta^d - \bdelta \right)
  + \frac{\rho}{2} \left\| \tbtheta^d - \bdelta \right\|_2^2,
\end{align}
}
where $\blambda$ is the Lagrangian multiplier, and $\rho > 0$ is a penalty
parameter for the augmented term.

ADMM \citep{boyd2011distributed} alternatively minimizes the augmented
Lagrangian function \eqref{aeq:prob1-al} over \textit{two} blocks of variables,
leading to an efficient operator splitting framework for nonlinear programs with
\textit{nonsmooth} objective function and \textit{equality} constraints.  Hence
ADMM solves the original problem \eqref{eq:prob0} when reformulated as problem
\eqref{aeq:prob1} with a sequence of easier sub-problems.  Specifically, ADMM
solves problem \eqref{eq:prob0} by alternatively minimizing \eqref{aeq:prob1-al}
over variables $\{{\btheta^c}, {\tbtheta^d} \} $, and
$\{ \boldsymbol{\delta}, \mathbf z \} $. This can be equivalently converted into
3 sub-problems over variables $\{{\btheta^c}, {\tbtheta^d}\}$,
$\boldsymbol{\delta}$ and $\mathbf z$, respectively.  ADMM decomposes the
optimization variables into two blocks and alternatively minimizes the augmented
Lagrangian function \eqref{aeq:prob1-al} in the following manner at any ADMM
iteration $t$:
{\small
\begin{align}
  \label{aeq:cont-bb}
    \left\{ \iter{\btheta^c}{t+1}, \iter{\tbtheta^d}{t+1} \right\}
    & = \argmin_{\btheta^c, \tbtheta^d} \mathcal{L} \left(
      \iter{\bz}{t}, \btheta^c, \tbtheta^d, \iter{\bdelta}{t},
      \iter{\blambda}{t}
    \right) \\
  \label{aeq:int-bb} \left\{ \iter{\bdelta}{t+1}, \iter{\bz}{t+1} \right\}
    & = \argmin_{\bdelta, \bz} \mathcal{L}\left(
      \bz, \iter{\btheta^c}{t+1}, \iter{\tbtheta^d}{t+1}, \bdelta,
      \iter{\blambda}{t}
    \right) \\
  \label{aeq:lambda-update} \iter{\blambda}{t+1} & =
    \iter{\blambda}{t} + \rho \left(
      \iter{\tbtheta^d}{t+1} - \iter{\bdelta}{t+1}
  \right).
\end{align}
}
Problem \eqref{aeq:cont-bb} can be simplified by removing constant terms to get
{\small
\begin{align}
  \nonumber
  \left\{ \iter{\btheta^c}{t+1}, \right.
    & \left. \iter{\tbtheta^d}{t+1} \right\} \\
  \label{aeq:cont-bb-1}
  & = \argmin_{\btheta^c, \tbtheta^d} \quad \widetilde{f} \left(
    \iter{\bz}{t},  \left\{\btheta^c, \tbtheta^d \right\}
    \right) + I_\mathcal{C}(\btheta^c) + I_{\widetilde{\mathcal{D}}}(\tbtheta^d) \\
  \nonumber
  & \quad \quad \quad \quad \quad \quad
    + \iter{\blambda}{t}{}^\top \left( \tbtheta^d - \iter{\bdelta}{t} \right)
    + \frac{\rho}{2} \left\| \tbtheta^d - \iter{\bdelta}{t} \right\|_2^2, \\
  \label{aeq:cont-bb-2}
  & = \argmin_{\btheta^c, \tbtheta^d} \quad \widetilde{f} \left(
    \iter{\bz}{t},  \left\{\btheta^c, \tbtheta^d \right\}
  \right) + I_\mathcal{C}(\btheta^c) + I_{\widetilde{\mathcal{D}}}(\tbtheta^d)
  + \frac{\rho}{2} \left\| \tbtheta^d - \mathbf{b} \right\|_2^2 \\
  \nonumber
  & \quad \quad \quad \quad \quad \quad
    \mbox{ where } \mathbf{b}
    = \iter{\bdelta}{t} - \frac{1}{\rho} \iter{\blambda}{t}.
\end{align}
}
A similar treatment to problem \eqref{aeq:int-bb} gives us
{\small
\begin{align}
  \label{aeq:int-bb-1} \left\{ \iter{\bdelta}{t+1}, \iter{\bz}{t+1} \right\}
    & = \argmin_{\bdelta, \bz} \quad \widetilde{f} \left(\bz, \left\{
      \iter{\btheta^c}{t+1}, \iter{\tbtheta^d}{t+1}
    \right\} \right) + I_\mathcal{Z}(\bz) \\
  \nonumber & \quad \quad \quad \quad \quad \quad
    + I_\mathcal{D}(\bdelta)
    + \iter{\blambda}{t}{}^\top \left( \iter{\tbtheta^d}{t+1} - \bdelta \right)
    + \frac{\rho}{2} \left\| \iter{\tbtheta^d}{t+1} - \bdelta \right\|_2^2, \\
  \label{aeq:int-bb-2} & = \argmin_{\bdelta, \bz} \quad
    \widetilde{f} \left(\bz,  \left\{
      \iter{\btheta^c}{t+1}, \iter{\tbtheta^d}{t+1}
    \right\} \right) + I_\mathcal{Z}(\bz) \\
  \nonumber & \quad \quad \quad \quad \quad
    + I_\mathcal{D}(\bdelta)
    + \frac{\rho}{2} \left\| \mathbf{a} - \bdelta \right\|_2^2
    \mbox{ where } \mathbf{a} =
    \iter{\tbtheta^d}{t+1} + \frac{1}{\rho} \iter{\blambda}{t}.
\end{align}
}
This simplification exposes the independence between $\bz$ and $\bdelta$,
allowing us to solve problem \eqref{aeq:int-bb} independently for $\bz$ and
$\bdelta$ as:
{\small
\begin{align}
  \label{aeq:int-bb-2d} \iter{\bdelta}{t+1} & = \argmin_{\bdelta} \quad
    I_\mathcal{D}(\bdelta) + \frac{\rho}{2}
    \left\| \mathbf{a} - \bdelta \right\|_2^2 \mbox{ where }
    \mathbf{a} = \iter{\tbtheta^d}{t+1} + \frac{1}{\rho} \iter{\blambda}{t}, \\
  \label{aeq:int-bb-2z} \iter{\bz}{t+1} & = \argmin_\bz \quad
    \widetilde{f} \left(\bz,  \left\{
      \iter{\btheta^c}{t+1}, \iter{\tbtheta^d}{t+1}
    \right\} \right) + I_\mathcal{Z}(\bz).
\end{align}
}
So we are able to decompose problem (3) into problems \eqref{aeq:cont-bb-2},
\eqref{aeq:int-bb-2d} and \eqref{aeq:int-bb-2z} which can be solved iteratively
along with the $\blambda^{(t)}$ updates (see Table 1).


\section{ADMM parameter sensitivity} \label{asec:ADMM-rho-sense}

\begin{figure}[htb]
  \centering
  \begin{subfigure}{0.4\textwidth}
    \includegraphics[width=\textwidth]{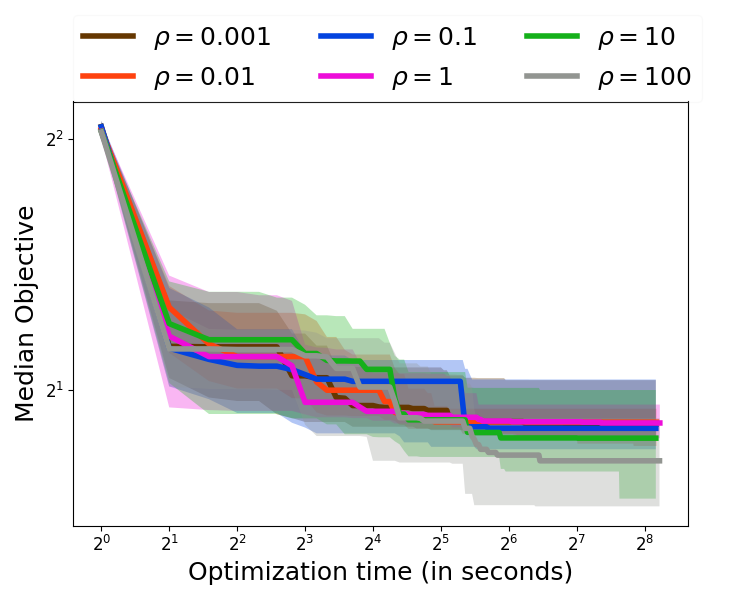}
    \caption{Run time in log scale}
  \end{subfigure}
  ~
  \begin{subfigure}{0.4\textwidth}
    \includegraphics[width=\textwidth]{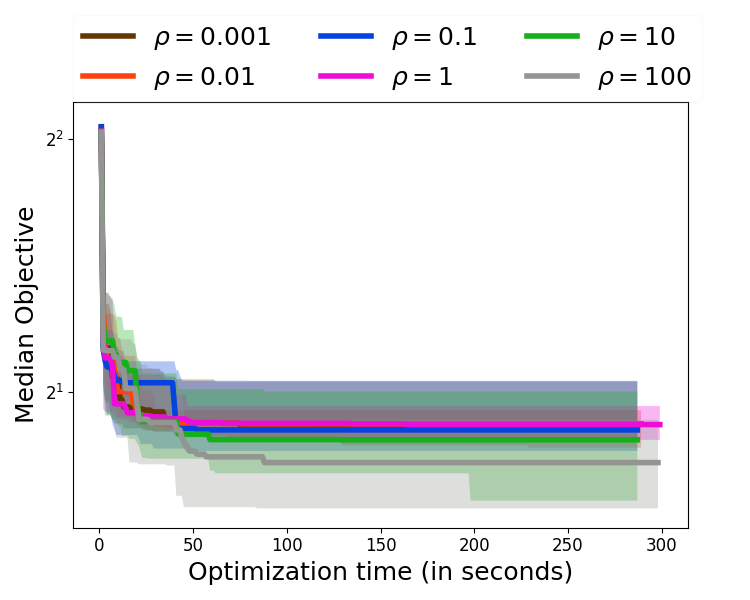}
    \caption{Run time in linear scale}
  \end{subfigure}
  \caption{ADMM {\bf sensitivity to the penalty parameter $\rho$} with adaptive
    ADMM. The performance is aggregated over 20 trials. The run time is
    dominated by the black-box optimization scheme, which, in this case, is a
    Bayesian optimization via Gaussian Process Regression.}
  \label{fig:abbobj-rhosense-adaptive}
\end{figure}

The performance of ADMM is dependent on the choice of the penalty parameter
$\rho > 0$ for the augmented term in the augmented Lagrangian
\eqref{aeq:prob1-al}. Figure \ref{fig:abbobj-rhosense-adaptive} shows the
dependence of adaptive precision ADMM on $\rho$ with the artificial black-box
objective (Appendix~\ref{asec:art-bbopt}). The results indicate that adaptive
ADMM for CASH is fairly robust to the choice of $\rho$.

\section{Empirical Evaluation of ADMM vs. JOPT on OpenML
  data} \label{asec:eval-admm-jopt} 

We compare the performance of the adaptive precision ADMM with different
sub-problem solvers to JOPT(BO) (BO on the joint problem with all the variables)
on 8 OpenML~\citep{bischl2017openml} binary classification data sets where we
optimize for the area under the ROC curve on 10\% of the data as the validation
set. Both adaptive precision ADMM and JOPT(BO) are executed for 1 hour with 10
repetitions, and the incumbent objective is averaged over the 10
repetitions. All evaluations were run single-threaded on a 8 core 8GB CentOS
virtual machines. For each data set, we note the objective $f_J$ achieved by
JOPT(BO) at the end of 1 hour and note the time $T_A$ (in seconds) required by
adaptive precision ADMM to reach that objective. Then we also note the best
objective achieved by ADMM at the end of 1 hour $f_A$. The speedup $S$ for ADMM
is computed as $S = \frac{3600}{T_A}$ and the improvement (in \%) $I = 100 *
\frac{f_J - f_A}{f_J}$.

We consider two versions of adaptive precision ADMM -- (i) using BO for both the
\eqref{eq:theta-min} and \eqref{eq:z-step} to get AdADMM(BO,BO), and (ii) using
BO for the \eqref{eq:theta-min} and Combinatorial multi-armed bandits (Ba) for
the \eqref{eq:z-step} to get AdADMM(BO,Ba). The improvements of ADMM over
JOPT(BO) are summarized in Table \ref{tab:opsplit-gains}, indicating significant
speedup (over $10\times$ in most cases) and further improvement (over $10\%$ in
many cases).
Table \ref{tab:opsplit-gains} shows that between AdADMM(BO,BO) and
AdADMM(BO,Ba), the latter provides {\em significantly higher speedups}, but the
former provides higher additional improvement in the final objective. This
demonstrates ADMM's flexibility, for example, allowing choice between faster or
more improved solution.

\begin{table}[t]
    \centering
    \begin{adjustbox}{max width=0.3\textwidth }
    \begin{tabular}{|l|r|r|r|r|}
        \hline
        Dataset & $\mathbf{S}_{\mbox{Ba}}$  & $\mathbf{S}_{\mbox{BO}}$ & $\mathbf{I}_{\mbox{Ba}}$ & $\mathbf{I}_{\mbox{BO}}$ \\
        \hline\hline
        Bank8FM    & $10 \times$ & $2 \times$ & $0 \%$ & $5 \%$\\
        CPU small  & $4  \times$ & $5 \times$ & $0 \%$ & $5 \%$\\
        fri-c2     & $153\times$ & $25\times$ & $56\%$ & $64\%$\\
        PC4        & $42 \times$ & $5 \times$ & $8 \%$ & $13\%$\\
        Pollen     & $25 \times$ & $7 \times$ & $4 \%$ & $3 \%$\\
        Puma8NH    & $11 \times$ & $4 \times$ & $1 \%$ & $1 \%$\\
        Sylvine    & $9  \times$ & $2 \times$ & $9 \%$ & $26\%$\\
        Wind       & $40 \times$ & $5 \times$ & $0 \%$ & $5 \%$\\
        \hline
    \end{tabular}
    \end{adjustbox}
    \caption{Comparing ADMM schemes to JOPT(BO), we list the speedup $\mathbf{S}_{\mbox{Ba}}$ \& $\mathbf{S}_{\mbox{BO}}$ achieved by AdADMM(BO,Ba) \& AdADMM(BO,BO) respectively to reach the best objective of JOPT, and the final objective improvement $\mathbf{I}_{\mbox{Ba}}$ \& $\mathbf{I}_{\mbox{BO}}$ (respectively) over the JOPT objective.
    These numbers are generated using the aggregate performance of JOPT and AdADMM over 10 trials.%
    }
    \label{tab:opsplit-gains}
\end{table}

\section{ADMM Decomposition for \eqref{eq:prob0} with black-box
  constraints} \label{asec:ADMM:csts}  

Without loss of generality, we assume that
$\epsilon_m \geq 0$ for $m \in [M]$.  By introducing scalars
$u_m \in [0, \epsilon_m]$, we can reformulate the inequality constraint
\eqref{eq:prob0-csts} as the equality constraint together with a box constraint
{\small
\begin{equation} \label{aeq:prob0-csts1}
  g_m\left(\bz, \left  \{ \btheta^c, \btheta^d \right \} \right)-
  \epsilon_m + u_m = 0, u_m \in [0, \epsilon_m],~ m \in [M].
\end{equation}
}
We then introduce a continuous surrogate black-box functions $\widetilde{g}_m$
for $g_m, \forall m \in [M]$ in a similar manner to $\widetilde{f}$ given by
\eqref{aeq:bbloss-continuous-surrogate}.  Following the reformulation of
\eqref{eq:prob0} that lends itself to the application of ADMM, the version
with black-box constraints \eqref{aeq:prob0-csts1} can be equivalently
transformed into
{\small
\begin{equation} \label{aeq:bb-csts-rr}
  \displaystyle \min_{\bz, \btheta^c, \tbtheta^d, \bdelta}
  \widetilde{f} \left(
    \bz, \left\{\btheta^c, \tbtheta^d \right\}
  \right) \st \left\{ \begin{array}{l}
    \bz_i \in \{0,1 \}^{K_i}, \mathbf 1^{\top} \bz_i = 1, \forall i \in [N] \\
    \btheta_{ij}^c \in \mathcal{C}_{ij}, \tbtheta_{ij}^d \in
    \widetilde{\mathcal{D}}_{ij}, \forall i \in [ N ], j \in [ K_i ] \\
    \bdelta_{ij} \in \mathcal{D}_{ij}, \forall i \in [ N ], j \in [ K_i ] \\
    \tbtheta_{ij}^d  = \bdelta_{ij}, \forall i \in [ N ], j \in [ K_i ] \\
    u_m \in [0, \epsilon_m], \forall m \in [M] \\
    \widetilde{g}_m \left(\bz,  \left\{\btheta^c, \tbtheta^d \right\}
    \right) -  \epsilon_m + u_m = 0, \forall m \in [M].
  \end{array} \right.
\end{equation}
}
Compared to problem \eqref{aeq:prob0-rr}, the introduction of auxiliary variables
$\{ u_m\}$ enables ADMM to incorporate \textit{black-box equality} constraints
as well as elementary \textit{white-box} constraints.

Defining $\mathcal{U} = \{ \mathbf{u} \colon \mathbf{u} = \{ u_m \in [0,
  \epsilon_m] \forall m \in [M] \} \}$, we can go through the mechanics of ADMM
to get the augmented Lagrangian with $\blambda$ and $\mu_m \forall m \in [M]$ as
the Lagrangian multipliers and $\rho > 0$ as the penalty parameter as follows:
{\small
\begin{equation*} \label{aeq:bbc-al}
  \begin{split}
    \mathcal{L} & \left( \bz, \btheta^c, \tbtheta^d, \bdelta, \bu, \blambda,
      \boldsymbol{\mu} \right) = \\
    & \widetilde{f} \left(
        \bz,  \left\{\btheta^c, \tbtheta^d \right\}
      \right) + I_\mathcal{Z}(\bz) + I_\mathcal{C}(\btheta^c)
      + I_{\widetilde{\mathcal{D}}}(\tbtheta^d) + I_\mathcal{D}(\bdelta)
      + \blambda^\top \left( \tbtheta^d - \bdelta \right)
      + \frac{\rho}{2} \left\| \tbtheta^d - \bdelta \right\|_2^2 \\
    & + I_\mathcal{U}(\bu) + \sum_{i=1}^{M} \mu_m \left( \widetilde{g}_m \left(
        \bz,  \left\{\btheta^c, \tbtheta^d \right\}
      \right) -  \epsilon_m + u_m \right) + \frac{\rho}{2} \sum_{i=1}^{M} \left(
      \widetilde{g}_m \left(
        \bz,  \left\{\btheta^c, \tbtheta^d \right\}
      \right) -  \epsilon_m + u_m  \right)^2.
  \end{split}
\end{equation*}
}
ADMM decomposes the optimization variables into two blocks for alternate
minimization of the augmented Lagrangian in the following manner at any ADMM
iteration $t$
{\small
\begin{align}
  \label{aeq:cbbc} \left\{ \iter{\btheta^c}{t+1}, \iter{\tbtheta^d}{t+1},
    \iter{\bu}{t+1}\right\}  & = \argmin_{\btheta^c, \tbtheta^d, \bu}
    \mathcal{L}\left( \iter{\bz}{t}, \btheta^c, \tbtheta^d, \iter{\bdelta}{t},
    \bu, \iter{\blambda}{t}, \iter{\boldsymbol{\mu}}{t} \right) \\
  \label{aeq:ibbc} \left\{ \iter{\bdelta}{t+1}, \iter{\bz}{t+1} \right\}
    & = \argmin_{\bdelta, \bz}  \mathcal{L}\left(
    \bz, \iter{\btheta^c}{t+1}, \iter{\tbtheta^d}{t+1}, \bdelta,
    \iter{\bu}{t+1}, \iter{\blambda}{t}, \iter{\boldsymbol{\mu}}{t} \right) \\
  \label{aeq:lup} \iter{\blambda}{t+1} & = \iter{\blambda}{t} + \rho \left(
    \iter{\tbtheta^d}{t+1} - \iter{\bdelta}{t+1} \right) \\
  \label{aeq:mup} \forall m \in [M],\ \  \iter{\mu_m}{t+1} & = \iter{\mu_m}{t} +
    \rho \left( \widetilde{g}_m ( \iter{\bz}{t+1},  \{\iter{\btheta^c}{t+1},
    \iter{\tbtheta^d}{t+1} \} ) -  \epsilon_m + \iter{u_m}{t+1}
    \right).
\end{align}
}
Note that, unlike the unconstrained case, the update of the augmented Lagrangian
multiplier $\mu_m$ requires the evaluation of the black-box function for the
constraint $g_m$. Simplifying problem \eqref{aeq:cbbc} gives us
{\small
\begin{equation} \label{aeq:cbbc-s-0}
  \begin{split}
    \min_{\btheta^c, \tbtheta^d, \bu} & \quad
      \widetilde{f} \left( \iter{\bz}{t},  \left\{\btheta^c, \tbtheta^d \right
      \} \right) \\
    & \quad \quad \quad + \frac{\rho}{2} \left[
      \left\| \tbtheta^d - \mathbf{b} \right\|_2^2 + \sum_{i=1}^{M} \left[
        \widetilde{g}_m \left( \iter{\bz}{t},  \left\{\btheta^c, \tbtheta^d
        \right \} \right) - \epsilon_m + u_m  +
        \frac{1}{\rho}\iter{\mu_m}{t} \right]^2 \right] \\
    & \st \left\{ \begin{array}{l}
        \btheta_{ij}^c \in \mathcal{C}_{ij} \forall i \in [N], j \in [K_i],\\
        \tbtheta_{ij}^d \in \widetilde{\mathcal{D}}_{ij} \forall i \in [N], j
        \in [K_i], \\
        u_m \in [0, \epsilon_m],
      \end{array} \right. \mbox{ where } \mathbf{b} = \iter{\bdelta}{t} -
      \frac{1}{\rho} \iter{\blambda}{t},
  \end{split}
\end{equation}
}
which can be further split into active and inactive set of continuous variables
based on the $\iter{\bz}{t}$ as in the solution of problem \eqref{aeq:cont-bb-2}
(the $\btheta$-min problem). The main difference from the unconstrained case in
problem \eqref{aeq:cont-bb-2} (the $\btheta$-min problem) to note here is that
the black-box optimization with continuous variables now has $M$ new variables
$u_m$ ($M$ is the total number of black-box constraints) which are active in
every ADMM iteration. This problem \eqref{aeq:cbbc-s-0} can be solved in the same
manner as problem \eqref{aeq:cont-bb-2} ($\btheta$-min) using SMBO or TR-DFO
techniques.

Simplifying and utilizing the independence of $\bz$ and $\bdelta$, we can split
problem \eqref{aeq:ibbc} into the following problem for $\bdelta$
{\small
\begin{equation} \label{aeq:ibbc-sd-0}
  \min_{  \bdelta  } \quad \frac{\rho}{2} \| \bdelta -  {\mathbf a} \|_2^2 \quad
    \st \bdelta_{ij} \in \mathcal D_{ij} \forall i \in [N], j \in [K_i]
    \mbox{ where } \mathbf{a} = \iter{\tbtheta^d}{t+1} + \frac{1}{\rho}
    \iter{\blambda}{t},
\end{equation}
}
which remains the same as problem \eqref{aeq:int-bb-2d} (the $\bdelta$-min
problem) in the unconstrained case, while the problem for $\bz$ becomes
{\small
\begin{equation} \label{aeq:ibbc-sz-0}
  \begin{split}
    \min_\bz & \quad \widetilde{f} (\bz,  \{\iter{\btheta^c}{t+1},
      \iter{\tbtheta^d}{t+1} \} ) \\
    & \quad \quad \quad + \frac{\rho}{2} \sum_{i=1}^{M} \left[
      \widetilde{g}_m ( \bz, \{\iter{\btheta^c}{t+1}, \iter{\tbtheta^d}{t+1} \}
     ) - \epsilon_m + \iter{u_m}{t+1}  + \frac{1}{\rho}
      \iter{\mu_m}{t}  \right]^2 \\
    & \st \bz_i \in \{0,1 \}^{K_i}, \mathbf 1^{\top} \bz_i = 1, \forall i \in
      [N].
  \end{split}
\end{equation}
}
The problem for $\bz$ is still a black-box integer programming problem, but now
with an updated black-box function and can be handled with techniques proposed
for the combinatorial problem  \eqref{aeq:int-bb-2z} in the absence of black-box
constraints (the $\bz$-min problem).


\end{document}